\documentclass[10pt,twocolumn,letterpaper]{article}

\usepackage{wacv}              

\definecolor{wacvblue}{rgb}{0.21,0.49,0.74}
\usepackage[pagebackref,breaklinks,colorlinks,allcolors=wacvblue]{hyperref}

\usepackage{multirow}
\usepackage{makecell}
\usepackage[utf8]{inputenc}
\usepackage[T1]{fontenc}

\def\wacvPaperID{994} 
\def\confName{WACV}
\def\confYear{2027}

\title{ComplexSync: High-Fidelity and Real-Time Lip Sync in Complex Scenarios}

\author{Jiaran Cai \quad \quad \quad \quad Xingpei Ma \quad \quad \quad \quad Shenneng Huang\\
Guangzhou Quwan Network Technology\\
\url{https://github.com/Playmate111/ComplexSync}
}

\begin{document}
\maketitle
\begin{abstract}
Lip synchronization aims to generate visual lip dynamics that align precisely with speech audio. Despite the high generation quality of diffusion models, they often struggle in complex scenarios and suffer from prohibitive inference latency, limiting real-world deployment. We present ComplexSync, a unified diffusion-based framework that enables real-time, high-fidelity lip sync under complex conditions. First, we introduce a dual-stream joint training strategy to mitigate information leakage from reference frames while preserving natural dynamics. Second, we develop a distillation-based acceleration scheme for single-step denoising, achieving a throughput of over 70 FPS. Third, we propose a relational alignment loss that leverages structural priors from Vision Foundation Models (VFMs) to enhance robustness against complex scene factors. Furthermore, we present the first benchmark specifically designed for complex lip synchronization, comprising over 200 challenging video sequences and specialized metrics. Extensive experiments demonstrate that ComplexSync achieves state-of-the-art performance across both standard and complex scenarios while enabling real-time inference.
\end{abstract}
    
\section{Introduction}
\label{sec:intro}
\vskip -0.1in
\begin{figure*}[!t]
\vskip -0.15in
\centering
\includegraphics[width=0.9\textwidth]{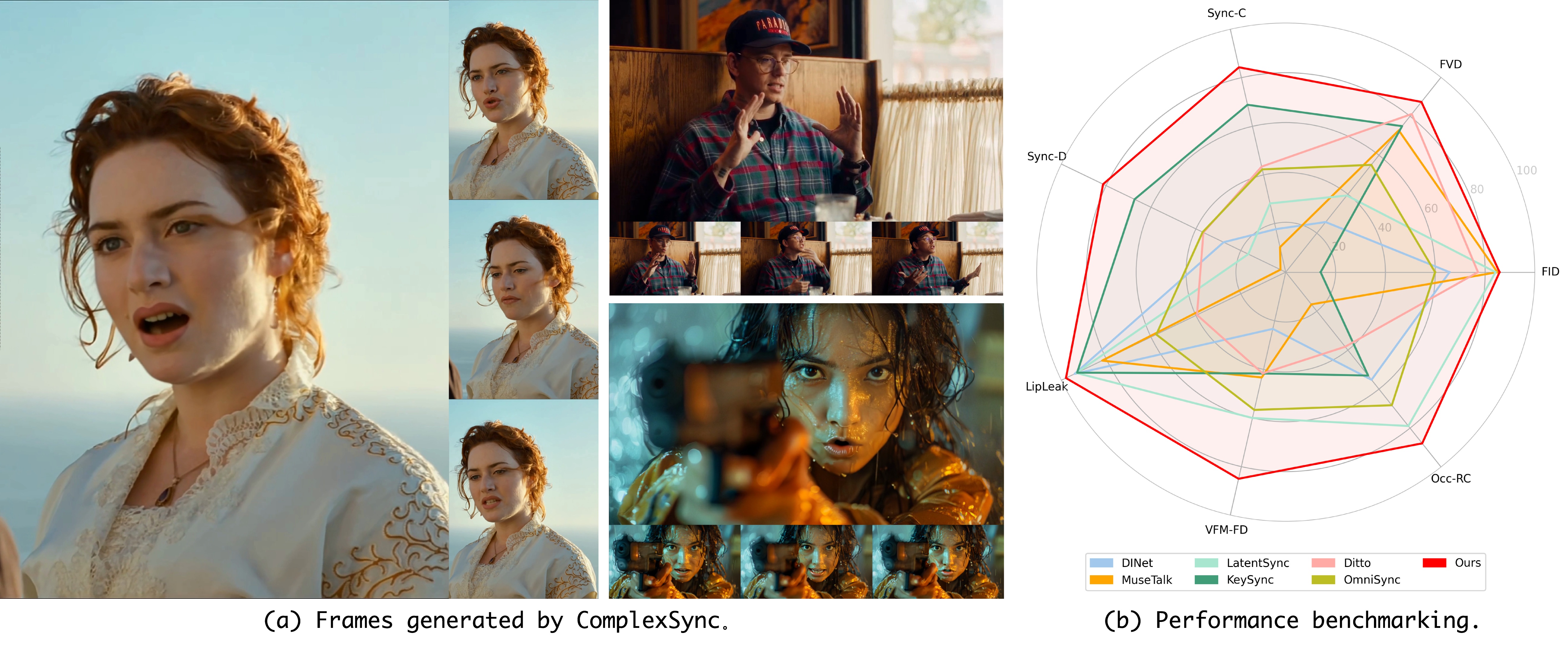}
\vskip -0.1in
\caption{We propose ComplexSync, a unified diffusion-based framework for real-time, high-fidelity lip synchronization in challenging scenarios. Extensive experiments show that ComplexSync achieves SOTA performance across both standard and challenging scenarios while supporting real-time inference.}
\label{fig:showcase}
\vskip -0.1in
\end{figure*}
Lip synchronization~\cite{xue2025human} is an audio-driven video-to-video task that generates speech-aligned facial dynamics while preserving identity and visual context. Driven by its transformative potential in applications such as film dubbing, digital avatars, and immersive virtual reality, this field has garnered significant research interest~\cite{prajwal2020lip,lu2021live,zhang2023dinet}. Recent advances in diffusion models~\cite{ho2020denoising,rombach2022high,podell2023sdxl} have elevated lip synchronization from a niche post-processing tool to a fundamental generative primitive. However, deploying these methods in real-world scenarios remains challenging. Specifically, achieving real-time inference alongside precise lip movements under unconstrained conditions, such as varying illumination, and facial occlusions, remains an open problem. 

Traditional GAN-based~\cite{goodfellow2020generative} approaches~\cite{prajwal2020lip,lu2021live,guan2023stylesync,zhang2023dinet,cheng2022videoretalking} primarily focus on photorealistic texture synthesis but often struggle with high-resolution generation and visual fidelity. More recently, diffusion-based methods~\cite{li2024latentsync,bigata2025keysync,ma2025sayanything,zhang2024musetalk} have emerged as a promising alternative, demonstrating strong cross-identity generalization without identity-specific fine-tuning. Nevertheless, key challenges remain. First, spatial information leakage remains unresolved, where the generated output inadvertently inherits unwanted motion or structural artifacts from the reference frames~\cite{muaz2023sidgan,yaman2024audio,cheng2022videoretalking,li2024latentsync}. Second, the high computational cost of the iterative denoising process leads to prohibitive inference latency, precluding real-time deployment. Finally, state-of-the-art (SOTA) methods exhibit limited robustness under complex conditions such as drastic illumination variations or partial facial occlusions, frequently resulting in visual artifacts or inaccurate lip movements. These shortcomings highlight a significant gap toward truly realistic and deployable lip synchronization. 

To bridge this gap, we propose ComplexSync, a diffusion-based framework for high-fidelity, real-time lip synchronization in unconstrained environments. 
First, we introduce a dual-stream joint training strategy: two functionally distinct yet architecturally identical experts are trained jointly with periodic weight fusion, effectively suppressing reference-inherited artifacts while preserving natural dynamics. Second, building on Score Distillation Sampling (SDS)~\cite{poole2022dreamfusion}, we develop a distillation-based acceleration scheme that enables single-step denoising, achieving over 70 FPS. Third, we introduce a relational alignment loss that leverages structural priors from Vision Foundation Models (VFMs)~\cite{radford2021learning,oquab2023dinov2,li2023blip,wang2023image} to implicitly encode complex scene factors, such as occlusions and illumination variations, thereby enhancing synchronization accuracy and realism under unconstrained conditions, particularly in cinematic and TV drama scenarios. 

Furthermore, we present the first benchmark specifically designed for complex lip synchronization, comprising over 200 challenging sequences and specialized metrics. Our contributions are:

\begin{itemize}
    \item A dual-stream joint training strategy that mitigates information leakage and enables high-fidelity lip-synced video generation. 
    \item A distillation-based acceleration scheme that achieves single-step denoising, enabling real-time inference at over 70 FPS. 
    \item A relational alignment loss leveraging VFMs to model scene complexity for robust lip movements under unconstrained conditions. 
    \item A new benchmark and specialized metrics for evaluating lip synchronization robustness in complex real-world scenarios.
\end{itemize}
\section{Related Work}
\label{sec:related_work}

\subsection{Audio-Driven Portrait Animation} 
Audio-driven portrait animation~\cite{jiang2024audio} is a broader task than lip synchronization that synthesizes lifelike talking-face videos from a single image and speech audio~\cite{vougioukas2019end,vougioukas2020realistic,tian2024emo,xu2024hallo,ma2025playmate}. Framed as an image-to-video task, it synthesizes entire frames end-to-end without constraining head pose or expression, bypassing complex blending in video-to-video pipelines. Early GAN-based methods~\cite{vougioukas2019end,vougioukas2020realistic,zhou2019talking,zhang2023sadtalker,zhou2021pose} often struggled to model cross-modal correlations between speech prosody and facial dynamics, leading to stiff and unexpressive results. Recent work has shifted to end-to-end diffusion models that map audio directly to motion, eliminating reliance on intermediate priors like 3DMMs or landmarks~\cite{tian2024emo,xu2024hallo,ma2025playmate,ji2025sonic}. Representative approaches like EMO~\cite{tian2024emo}, Hallo~\cite{xu2024hallo}, and EchoMimic~\cite{chen2025echomimic} leverage the generative capabilities of Stable Diffusion~\cite{rombach2022high} to produce temporally coherent animations. More recently, Diffusion Transformers (DiTs)~\cite{peebles2023scalable} have emerged as a powerful backbone for this task. Methods such as OmniHuman-1~\cite{lin2025omnihuman} and InfiniteTalk~\cite{yang2025infinitetalk} leverage large-scale DiT architectures~\cite{seawead2025seaweed,wan2025wan} to extend animation to the upper body or full body, achieving remarkable temporal smoothness~\cite{wang2025fantasytalking,kong2025let,ma2025playmate2}. However, these holistic models remain suboptimal for dedicated lip sync due to shortcut learning, as observed in LatentSync~\cite{li2024latentsync}.

\begin{figure*}[!t]
  \vskip -0.15in
\centering
\includegraphics[width=0.9\textwidth]{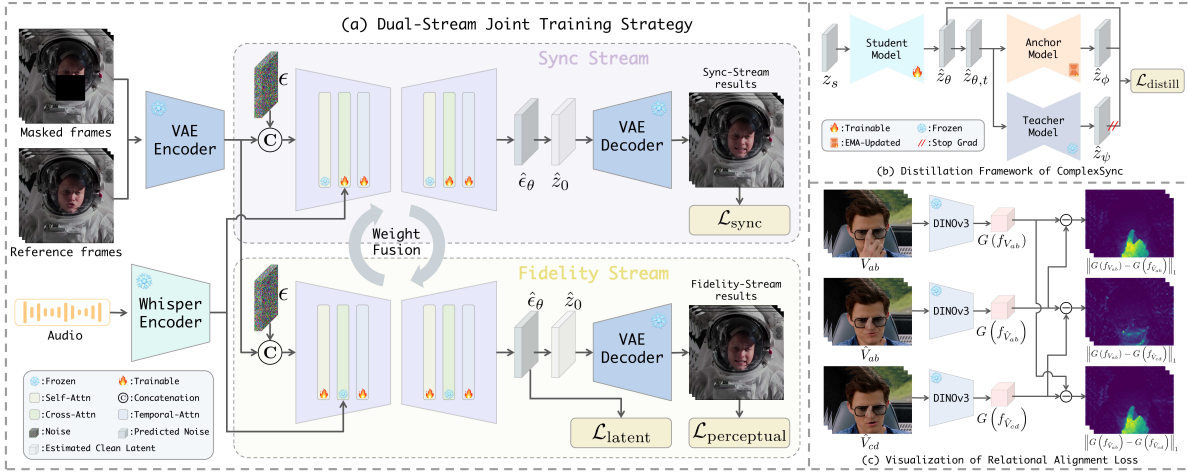}
\caption{Overview of ComplexSync. (a) Dual-stream joint training strategy designed to suppress spatial information leakage and ensure high-fidelity synthesis. (b) Distillation-based acceleration scheme that facilitates single-step denoising for real-time inference. (c) Relational alignment loss leveraging VFMs to enhance synchronization precision and robustness under unconstrained conditions.}
\label{fig:framework}
\vskip -0.1in
\end{figure*}

\subsection{Audio-Driven Lip Synchronization}
Unlike portrait animation, audio-driven lip synchronization~\cite{zhong2023identity,wang2023seeing,mukhopadhyay2024diff2lip,muaz2023sidgan,yaman2024audio,yaman2024audio2,wang2023lipformer,peng2024synctalk,park2022synctalkface,shen2023difftalk,stypulkowski2024diffused,cheng2022videoretalking,zhang2024musetalk} is a video-to-video editing task that necessitates preserving the original visual context while modifying only the mouth dynamics. Seminal works like Wav2Lip~\cite{prajwal2020lip} pioneered audio-visual alignment via pre-trained SyncNet~\cite{chung2016out}. Subsequent research has leveraged StyleGAN2-based architectures~\cite{karras2020analyzing} to bolster identity consistency (e.g., StyleSync~\cite{guan2023stylesync}, StyleLipSync~\cite{ki2023stylelipsync}) or capitalized on 3D mesh representations~\cite{guan2024resyncer} to achieve higher geometric fidelity. Alternatively, DINet~\cite{zhang2023dinet} introduced spatial feature deformation as a mechanism to synthesize precise mouth articulations. Recently, diffusion models have driven significant progress: LatentSync~\cite{li2024latentsync} introduced an end-to-end latent diffusion framework; KeySync~\cite{bigata2025keysync} proposed a two-stage pipeline with a tailored masking strategy to mitigate expression leakage and handle facial occlusions; and OmniSync~\cite{pengomnisync} developed a mask-free DiT architecture with dynamic spatiotemporal guidance to suppress lip shape leakage. Despite these advances, critical bottlenecks persist. First, the iterative nature of diffusion models leads to prohibitive inference latency, precluding real-time deployment~\cite{li2024latentsync}. Second, existing methods exhibit limited environmental robustness and often struggle in unconstrained real-world settings characterized by physical occlusions and volatile illumination variations. In contrast, our method, ComplexSync, directly addresses both challenges by enabling real-time, high-fidelity lip synchronization in complex, real-world environments. 

\subsection{Diffusion Distillation Models}
Accelerating inference is a pivotal challenge in diffusion research~\cite{shen2025efficient}. Current strategies primarily bifurcate into training-free caching~\cite{liu2025timestep,zou2024accelerating,selvaraju2024fora} and training-intensive distillation~\cite{huang2025self,luo2023latent,yin2024improved,sauer2024adversarial}. While caching methods like TeaCache~\cite{liu2025timestep} offer engineering-level speedups, distillation paradigms such as LCM~\cite{luo2023latent} and ADD~\cite{sauer2024adversarial} provide more substantial theoretical acceleration. In the animation domain, recent works have also explored efficiency: AniTalker~\cite{liu2024anitalker} and Ditto~\cite{li2025ditto} employ lightweight diffusion models for motion prediction; autoregressive-based methods~\cite{chu2025artalk} enable streaming via context-aware generation; and READ~\cite{wang2025read} achieves real-time talking-head synthesis in a highly compressed latent space. However, achieving high-fidelity, real-time lip synchronization via diffusion remains largely underexplored. Motivated by this, we adapt and extend the SDS framework to facilitate single-step denoising, enabling a significant throughput of over 70 FPS. 
\section{Methodology}
In this section, we present ComplexSync, a unified diffusion-based framework for real-time, high-fidelity lip synchronization in complex scenarios (overview in \cref{fig:framework}). Our approach comprises three core components: (1) a dual-stream joint training strategy (\cref{subsec:3_1}) that mitigates reference-frame information leakage; (2) a single-step denoising paradigm (\cref{subsec:3_2}) based on score distillation for real-time inference; and (3) a relational alignment loss (\cref{subsec:3_3}) that enhances robustness to challenging scene factors such as volatile illumination and facial occlusions. 

\subsection{Dual-stream Joint Training Strategy}
\label{subsec:3_1}
\textbf{Motivation and Dual-Stream Strategy.} A primary challenge in diffusion-based lip synchronization is spatial information leakage~\cite{li2024latentsync,bigata2025keysync}, where the model over-prioritizes structural cues from reference frames at the expense of audio conditioning, thereby inducing spurious facial motion artifacts. While recent methods (e.g., LatentSync~\cite{li2024latentsync}, KeySync~\cite{bigata2025keysync}) attempt to mitigate this, they often struggle in complex scenarios. To address this, we propose a dual-stream joint training strategy comprising a synchronization stream optimized for precise audio-visual alignment and a fidelity stream dedicated to identity and background consistency.  

\textbf{Task Definition and Backbone.} Let $V$ denote the ground-truth video. Our framework generates $\hat{V}$ conditioned on speech $A$ and reference frames $V_{\text{ref}}$. As shown in \cref{fig:framework}(a), we build upon Stable Diffusion (initialized from SD 1.5) with a pre-trained Whisper~\cite{radford2023robust} audio encoder. Audio embeddings are integrated into the U-Net via cross-attention. In addition, we incorporate a temporal attention module to enhance inter-frame consistency. In contrast to the fixed two-stage training in~\cite{li2024latentsync,bigata2025keysync}, we conceptualize the temporal attention mechanism as a pluggable temporal smoother. During training, this module is stochastically bypassed following a CFG-like dropout protocol. This strategy facilitates precise control over the degree of temporal stability during inference; by modulating the CFG scale, we can adaptively tune the smoothing intensity to balance motion vividness and temporal coherence. 

\textbf{Synchronization Stream.} This stream aims for precise audio-visual alignment by leveraging a pre-trained StableSyncNet~\cite{li2024latentsync} as a pixel-space supervisor. To bypass iterative sampling during training, we estimate the clean latent $\hat{z}_0$ in a single step from the predicted noise $\epsilon_\theta(z_t)$, following DDIM~\cite{song2020denoising}:
\vskip -0.11in
\begin{equation}
\resizebox{0.7\linewidth}{!}{
$\begin{aligned}
\hat z_0=\frac{z_t - \sqrt{1 - \overline \alpha _t} \epsilon_\theta \left( z_t,\mathcal{E} \left( V_{\text{ref}} \right) ,\mathcal{A} \left( A \right) ,t \right) }{\sqrt{\overline \alpha _t}},
\end{aligned}$%
}
\end{equation} 
where $A$ and $\mathcal{A}$ denote the input audio and the audio encoder, respectively; $\epsilon_\theta$ is the predicted noise parameterized by $\theta$; and $V_{\text{ref}}$ and $\mathcal{E}$ represent the reference frames and the VAE~\cite{kingma2013auto} encoder, respectively. To mitigate information leakage, we randomly select a single reference frame as $V_{\text{ref}}$ during this stage.
Only the audio cross-attention layers and temporal attention module are optimized; the rest of the U-Net is frozen. Given 16 decoded frames $\mathcal{D}(\hat{z}_0)_{i:i+16}$ and their corresponding audio segment $A_{i:i+16}$, the sync loss is formulated as:
\begin{equation}
{\mathcal{L}_{\text{sync}}}=\mathbb{E}_{z_0,A,t,\epsilon,i} \left[ \text{StableSyncNet} \left( \mathcal{D} \left( \hat{z}_0 \right) _{i:i+16},A_{i:i+16} \right) \right],
\end{equation}
where $\mathcal{D}$ represents the VAE decoder. 

\textbf{Fidelity Stream.} To preserve identity consistency and background integrity, this stream optimizes all parameters except the audio cross-attention layers. This ensures that the model focuses on spatial-temporal consistency without being adversely influenced by acoustic features. The objective combines a latent-space reconstruction loss and a pixel-space perceptual loss. The former is:
\begin{equation}
{\mathcal{L}_{\text{latent}}}=\mathbb{E}_{A, V_{\text{ref}}, t, \epsilon} \left[ \left\| \epsilon-\epsilon_\theta \left( z_t,\mathcal{E} \left( V_{\text{ref}} \right) ,\mathcal{A} \left( A \right) ,t \right) \right\|^{2}_{2} \right],
\end{equation}
where $\epsilon$ is additive Gaussian noise. To further refine visual fidelity, we employ a VGG-based~\cite{simonyan2014very} perceptual loss:
\begin{equation}
\mathcal{L}_{\text{perceptual}}=\mathbb{E}_{z_0,x,\epsilon,t,i} \left[ \left\| \mathcal{V} \left( \mathcal{D} \left( \hat{z}_0 \right) _i \right) - \mathcal{V} \left( x_i \right) \right\|_1 \right] ,
\end{equation} 
where $\mathcal{V}$ denotes the VGG-19 feature extractor and $i$ denotes the frame index. 
The joint objective is:
\begin{equation}
\mathcal{L}_{\text{total}}=\lambda_1 \mathcal{L}_{\text{latent}} + \lambda_2 \mathcal{L}_{\text{perceptual}},
\end{equation} 
where $\lambda_1 = 1.0$ and $\lambda_2 = 1.5$. 

\textbf{Joint Training Strategy.} In contrast to the conventional two-stage training used in methods like KeySync and LatentSync, we propose a unified joint training strategy that enables faster convergence and effectively suppresses information leakage. Specifically, upon establishing the two aforementioned streams, both branches are optimized in parallel. To facilitate synergistic learning, we introduce a periodic weight fusion mechanism. After a predefined number of iterations, the parameters of both expert branches are averaged and updated with their fused weights. The rationale is that each branch serves as a domain-specific expert—optimizing for either lip synchronization or identity preservation. By periodically averaging weights, each expert internalizes complementary representations from its counterpart, fostering multi-modal proficiency. Furthermore, this fusion mechanism ensures a smooth optimization trajectory, effectively mitigating catastrophic forgetting and steering the model toward a stable global optimum.

\subsection{Real-time Single-step Denoising}
\label{subsec:3_2}
In this section, we delineate the overall distillation framework of ComplexSync, designed to enable real-time inference while preserving the generative prior of the base model. As illustrated in~\cref{fig:framework}(b), our pipeline comprises three networks: a student, a teacher, and an anchor. All three are initialized with the pre-trained weights obtained in \cref{subsec:3_1}, and the teacher’s parameters are kept frozen throughout training. Notably, the teacher uses CFG during inference, while the student and anchor operate without it. 

Following SDS~\cite{poole2022dreamfusion}, we perform distillation in the latent space. For brevity, conditioning variables ($V_{\text{ref}}$, $A$) are omitted in the following. Given an input video $V$, we encode its latent $z_0$ via the VAE encoder and obtain noisy latents $z_s = \alpha_s z_0 + \sigma_s \epsilon$ through forward diffusion. The timestep $s$ is uniformly sampled from a discrete set $T_{\text{student}} = \{\tau_1, \dots, \tau_n\}$. In practice, we adopt single-step distillation with $N=1$ and fixed $\tau_1 = 1000$. Inspired by ADD~\cite{sauer2024adversarial}, we perturb the student’s output $\hat{z}_\theta(z_s, s)$ via forward diffusion to obtain $\hat{z}_{\theta,t} = \alpha_t \hat{z}_\theta(z_s, s) + \sigma_t \epsilon'$. This perturbed latent serves as input to both teacher and anchor models. Their denoising predictions, $\hat{z}_{\psi}(\hat{z}_{\theta,t}, t)$ and $\hat{z}_{\phi}(\hat{z}_{\theta,t}, t)$, act as distillation targets, supervising the student through a latent-space consistency loss. 

We define the distance function as $\operatorname{d}(x, y):=\|x-y\|_{2}^{2}$, yielding the original SDS loss:
\begin{equation}
\begin{split}
\mathcal{L}_{\text {SDS}}&= \mathbb{E}_{z_0, \epsilon, \epsilon', s, t} \left[ w \left( t \right) d \left( \hat{z}_{\theta} \left( z_s,s \right) , \hat{z}_{\psi} \left( \operatorname{sg} \left( \hat{z}_{\theta, t} \right), t \right) \right) \right] \\
&=\mathbb{E}_{z_0, \epsilon, \epsilon', s, t}\left[w \left( t \right) \left\| \hat{z}_{\theta} \left( z_s,s \right) - \hat{z}_{\psi} \left( \operatorname{sg} \left( \hat{z}_{\theta, t} \right), t \right) \right\|^{2}_{2} \right],
\end{split}
\end{equation}
where $\operatorname{sg}(\cdot)$ denotes stop-gradient. We employ the weighting factor $w(t) = \alpha_t$ to de-emphasize high-noise components, thereby prioritizing the refinement of clean structural details. 

Intuitively, SDS optimization aligns the student’s generated distribution with the teacher’s score manifold by minimizing their score discrepancy. However, a critical challenge arises: the teacher receives pseudo-latent samples derived from the student’s current output, which are then aggressively amplified by CFG. This dependency distorts the distillation target and often leads to mode collapse. In severe cases, it causes information leakage, where the student overfits to artifacts in the corrupted guidance signal, thereby compromising the fidelity and generalization of the distilled knowledge. 

To mitigate instability and objective distortion, we propose an enhanced SDS loss that improves convergence and training stability. Our approach introduces an anchor model updated via exponential moving average (EMA) to serve as a stable reference. Instead of supervising the student with the teacher alone, we use a joint prediction from both the teacher and the anchor as the distillation target. This design acts as a structural regularizer that shields the student from noisy gradients caused by the teacher's hallucinated latents under high CFG scales. Our final distillation objective is formulated as:
\begin{equation}
\resizebox{0.9\linewidth}{!}{
$\begin{aligned}
\mathcal{L}_{\text {distill}}&=\mathbb{E} \left[ w \left( t \right) \left\| \hat{z}_{\psi} \left( \operatorname{sg} \left( \hat{z}_{\theta, t} \right) , t \right) - \hat{z}_{\phi} \left( \hat{z}_{\theta, t}, t \right) \right\|^{2}_{2} \right] \\
&=\mathbb{E} \left[ \sigma_t \left( \hat{z}_{\psi} \left( \operatorname{sg} \left( \hat{z}_{\theta, t} \right) , t \right) - \hat{z}_{\phi} \left( \hat{z}_{\theta, t}, t \right) \right) ^{T} \left( \hat{z}_{\psi} \left( \operatorname{sg} \left( \hat{z}_{\theta, t} \right) , t \right) - \hat{z}_{\theta} \left( z_s,s \right) \right) \right],
\end{aligned}$%
}
\label{eq:loss_distill}
\end{equation}
where the expectation is taken over $\{z_0, \epsilon, \epsilon', s, t\}$. 

The derivation of \cref{eq:loss_distill} proceeds as follows. Under the diffusion framework and without loss of generality, we set $\alpha_t = 1$, so that $z_t = z_0 + \sigma_t \epsilon$ and $q \left( z_t \mid z_0 \right) = \mathcal{N}\left(z_0, \sigma_t^2 \mathbf{I} \right)$. This yields:
\vskip -0.1in
\begin{equation}
    q \left( z_t \mid z_0 \right) \propto \exp\!\left( -\frac{1}{2\sigma_t^2} \|z_t - z_0\|^2 \right),
    \label{eq:derivation_1}
\end{equation}
\vskip -0.1in
\begin{equation}
    \nabla_{z_t} \log q \left( z_t \mid z_0 \right) = -\frac{z_t - z_0}{\sigma_t^2},
    \label{eq:derivation_2}
\end{equation}
\vskip -0.1in
\begin{equation}
\begin{split}
\nabla_{z_t} q \left( z_t \mid z_0 \right) &= q \left( z_t \mid z_0 \right) \nabla_{z_t} \log q \left( z_t \mid z_0 \right) \\
&= -q \left( z_t \mid z_0 \right) \frac{z_t - z_0}{\sigma_t^2}.
\end{split}
\label{eq:derivation_3}
\end{equation}
\vskip -0.1in
The marginal distribution of the noisy data is $p \left( z_t \right) = \int q \left( z_t \mid z_0 \right) p \left( z_0 \right) \, dz_0$, with gradient $\nabla_{z_t} p \left( z_t \right) = \int \nabla_{z_t} q \left( z_t \mid z_0 \right) p \left( z_0 \right) \, dz_0$. Substituting \cref{eq:derivation_3} yields:
\begin{equation}
\resizebox{0.9\linewidth}{!}{
$\begin{aligned}
\nabla_{z_t} \log p \left( z_t \right) &= \int \frac{-q \left( z_t \mid z_0 \right) \frac{z_t - z_0}{\sigma_t^2}}{p \left( z_t \right)} p \left( z_0 \right) \, dz_0 = -\frac{1}{\sigma_t^2} \left( z_t - \frac{ \int z_0 q \left( z_t \mid z_0 \right) p \left( z_0 \right) \, dz_0 }{ p \left( z_t \right) } \right).
\end{aligned}$%
}
\label{eq:derivation_4}
\end{equation}
The posterior expectation of $z_0$ given $z_t$ is:
\begin{equation}
\resizebox{0.85\linewidth}{!}{
$\begin{aligned}
    \mathbb{E}\left( z_0 \mid z_t \right) 
    = \int z_0 \, q\left( z_0 \mid z_t \right) \, dz_0 
    = \frac{ \int z_0 \, q\left( z_t \mid z_0 \right) p\left( z_0 \right) \, dz_0 }{ p\left( z_t \right) }.
\end{aligned}$%
}
\label{eq:derivation_5}
\end{equation}
Combining \cref{eq:derivation_4} and \cref{eq:derivation_5}, we obtain:
\begin{equation}
    \nabla_{z_t} \log p\left( z_t \right) 
    = -\frac{1}{\sigma_t^2} \left( z_t - \mathbb{E}\left( z_0 \mid z_t \right) \right).
    \label{eq:derivation_6}
\end{equation}
By the Score Projection Identity~\cite{zhou2024score}, it holds that:
\begin{equation}
\resizebox{0.9\linewidth}{!}{
$\begin{aligned}
    \mathbb{E}_{z_t \sim p_\theta\left( z_t \right)} \left[ u^\top\!\left( z_t \right) \nabla_{z_t} \log p\left( z_t \right) \right] 
    = \mathbb{E}_{\left( z_t, z_0 \right) \sim q\left( z_t \mid z_0 \right) p_\theta\left( z_0 \right)} \left[ u^\top\!\left( z_t \right) \nabla_{z_t} \log q\left( z_t \mid z_0 \right) \right].
\end{aligned}$%
}
\label{eq:derivation_7}
\end{equation}
Combining \cref{eq:derivation_6} and \cref{eq:derivation_7}, we arrive at the following derivation for the distillation loss $\mathcal{L}_{\text{distill}}$:
\begin{equation}
\resizebox{0.98\linewidth}{!}{
$\begin{aligned}
\mathcal{L}&_{\text{distill}} 
= \mathbb{E}\left[ w\left( t \right) \left\| \hat{z}_{\psi}\left( \operatorname{sg}\left( \hat{z}_{\theta, t} \right), t \right) - \hat{z}_{\phi}\left( \hat{z}_{\theta, t}, t \right) \right\|^{2}_{2} \right] \\
&= \mathbb{E}\left[ w\left( t \right) \sigma_t^4 \left\| \nabla_{z_t} \log p_{T}\left( z_t \right) - \nabla_{z_t} \log p_{A}\left( z_t \right) \right\|_2^2 \right] \\
&= \mathbb{E}\left[ w\left( t \right) \sigma_t^4 \left( \nabla_{z_t} \log p_{T}\left( z_t \right) - \nabla_{z_t} \log p_{A}\left( z_t \right) \right)^{\!\top} \left( \nabla_{z_t} \log p_{T}\left( z_t \right) - \nabla_{z_t} \log p_{A}\left( z_t \right) \right) \right] \\
&= \mathbb{E}\left[ w\left( t \right) \sigma_t^4 \left( \nabla_{z_t} \log p_{T}\left( z_t \right) - \nabla_{z_t} \log p_{A}\left( z_t \right) \right)^{\!\top} \left( \nabla_{z_t} \log p_{T}\left( z_t \right) - \nabla_{z_t} \log q_{A}\left( z_t \mid z_0 \right) \right) \right] \\
&= \mathbb{E}\left[ w\left( t \right) \left( \hat{z}_{\psi}\left( \operatorname{sg}\left( \hat{z}_{\theta, t} \right), t \right) - \hat{z}_{\phi}\left( \hat{z}_{\theta, t}, t \right) \right)^{\!\top} \left( \hat{z}_{\psi}\left( \operatorname{sg}\left( \hat{z}_{\theta, t} \right), t \right) - \hat{z}_{\theta}\left( z_s, s \right) \right) \right].
\end{aligned}$%
}
\label{eq:loss_distill_final}
\end{equation}

\subsection{Relational Alignment Loss}
\label{subsec:3_3}
Despite achieving precise lip synchronization, diffusion-based generators often degrade in unconstrained scenarios involving occlusions, extreme lighting variations, or dynamic camera motion. To improve generation fidelity, we propose a relational alignment loss that leverages structural priors from VFMs to implicitly encode complex scene factors. Our approach is motivated by the insight that stronger perceptual grounding leads to higher-quality synthesis. Building on REPA~\cite{yu2024representation,leng2025repa,tian2025u}, which bridges pre-trained foundation models and generative diffusion models, and inspired by VideoREPA~\cite{zhang2025videorepa}, we employ DINOv3~\cite{simeoni2025dinov3} as our VFM backbone and inject scene-aware knowledge (e.g., occlusion patterns and illumination dynamics) into the lip-sync model via targeted fine-tuning. 

As shown in~\cref{fig:framework}(c), we employ a frozen VFM encoder $\mathcal{E}_v$ to extract spatiotemporal features $f_v \in \mathbb{R}^{f \times N \times D}$ from a video $V \in \mathbb{R}^{F \times C \times H \times W}$, where $N = h \times w$ is the number of tokens and $(F/f, H/h, W/w)$ denote the temporal and spatial compression ratios. Specifically, we sample a video clip $V_{ab}$ and two distinct audio segments $A_{ab}$ and $A_{cd}$, indexed by non-overlapping intervals $(a\!:\!b)$ and $(c\!:\!d)$. Treating $V_{ab}$ as the reference, we condition the generator on $A_{ab}$ and $A_{cd}$ to synthesize $\hat{V}_{ab}$ and $\hat{V}_{cd}$, respectively. The resulting features $f_{V_{ab}}$, $f_{\hat{V}_{ab}}$, and $f_{\hat{V}_{cd}}$ are then obtained via the same frozen $\mathcal{E}_v$. The relational alignment loss is defined as:
\begin{equation}
\resizebox{0.88\linewidth}{!}{
$\begin{aligned}
\mathcal{L}_{\text{RA}} = \left\| G \left( f_{V_{ab}} \right) - G \left( f_{\hat{V}_{ab}} \right) \right\|_1 + \left\| G \left( f_{V_{ab}} \right) - G \left( f_{\hat{V}_{cd}} \right) \right\|_1 - \left\| G \left( f_{\hat{V}_{ab}} \right) - G \left( f_{\hat{V}_{cd}} \right) \right\|_1.
\end{aligned}$%
}
\label{eq:loss_ra}
\end{equation}
Here, $G(f) = \text{Concat}(S_1, S_2, \dots, S_f) \in \mathbb{R}^{f \times N \times N}$, where each $S_t \in \mathbb{R}^{N \times N}$ is a spatial token-wise similarity matrix with entries
\begin{equation}
    S_{t,i,j} = \frac{f_{t,i}^\top f_{t,j}}{\|f_{t,i}\| \, \|f_{t,j}\|},
\end{equation}
denoting the cosine similarity between features at spatial positions $i$ and $j$ in frame $t$. We adopt this Gram-like representation because it is invariant to spatial layout, enabling robust capture of second-order statistics—such as texture, occlusion patterns, and illumination characteristics—while discarding positional information. 

\begin{table*}[!t]
    \vskip -0.15in
    \centering
    \belowrulesep=0pt
    \aboverulesep=0pt
    {\vskip -1.5mm}
    \renewcommand{\arraystretch}{1.3}
    \scalebox{0.85}{
    \begin{tabular}{l|cccccc|cccc|c}
    \toprule
    \multirow{2}{*}{Methods} & \multicolumn{6}{c|}{HDTF} & \multicolumn{4}{c|}{ComplexSync-Val} & \multirow{2}{*}{FPS$\uparrow$ } \\
    \cline{2-6} \cline{7-10}
    ~ & FID$\downarrow$ & FVD$\downarrow$ & CSIM$\uparrow$ & Sync-C$\uparrow$ & Sync-D$\downarrow$ & LipLeak$\downarrow$ & Sync-C$\uparrow$ & Sync-D$\downarrow$ & VFM-FD$\downarrow$ & Occ-RC$\uparrow$ \\
    \hline
    DINet & 10.14 & 59.72 & 0.835 & 4.62 & 9.17 & \underline{0.0596} & 3.52 & 7.69 & 0.086 & 87.70 & \underline{57.74} \\
    Musetalk & \underline{7.26} & 40.90 & 0.909 & 4.36 & 9.93 & 0.1822 & 2.83 & 9.60 & 0.074 & 78.80 & 20.16 \\
    LatentSync & 7.37 & 54.27 & 0.900 & 4.99 & 9.50 & 0.0685 & \underline{6.73} & \underline{7.37} & \underline{0.064} & \underline{93.10} & 2.13 \\
    KeySync & 17.90 & 40.04 & 0.801 & \underline{6.41} & \underline{7.98} & 0.0708 & 1.67 & 9.81 & 0.075 & 87.20 & 1.14 \\
    Ditto & 8.43 & \underline{37.55} & \underline{0.916} & 5.52 & 8.90 & 0.6048 & 4.26 & 9.61 & 0.075 & 83.80 & 16.6 \\
    OmniSync & 11.01 & 47.98 & 0.826 & 5.48 & 8.89 & 0.4236 & 3.86 & 10.55 & 0.066 & 90.70 & - \\
    \hline
    Ours & \textbf{7.13} & \textbf{35.07} & \textbf{0.927} & \textbf{6.95} & \textbf{7.56} & \textbf{0.0213} & \textbf{8.87} & \textbf{6.28} & \textbf{0.049} & \textbf{95.20} & \textbf{72.73} \\
    \bottomrule
    \end{tabular}
    }
    \vskip -0.08in
    \caption{Quantitative comparisons with existing methods on two test datasets. The best results are in \textbf{bold}, and the second-best are in \underline{underlined}.}
    \label{tab:quantitative_comparison}
    \vskip -0.15in
\end{table*}

\begin{figure}[!t]
\centering
\includegraphics[width=0.48\textwidth]{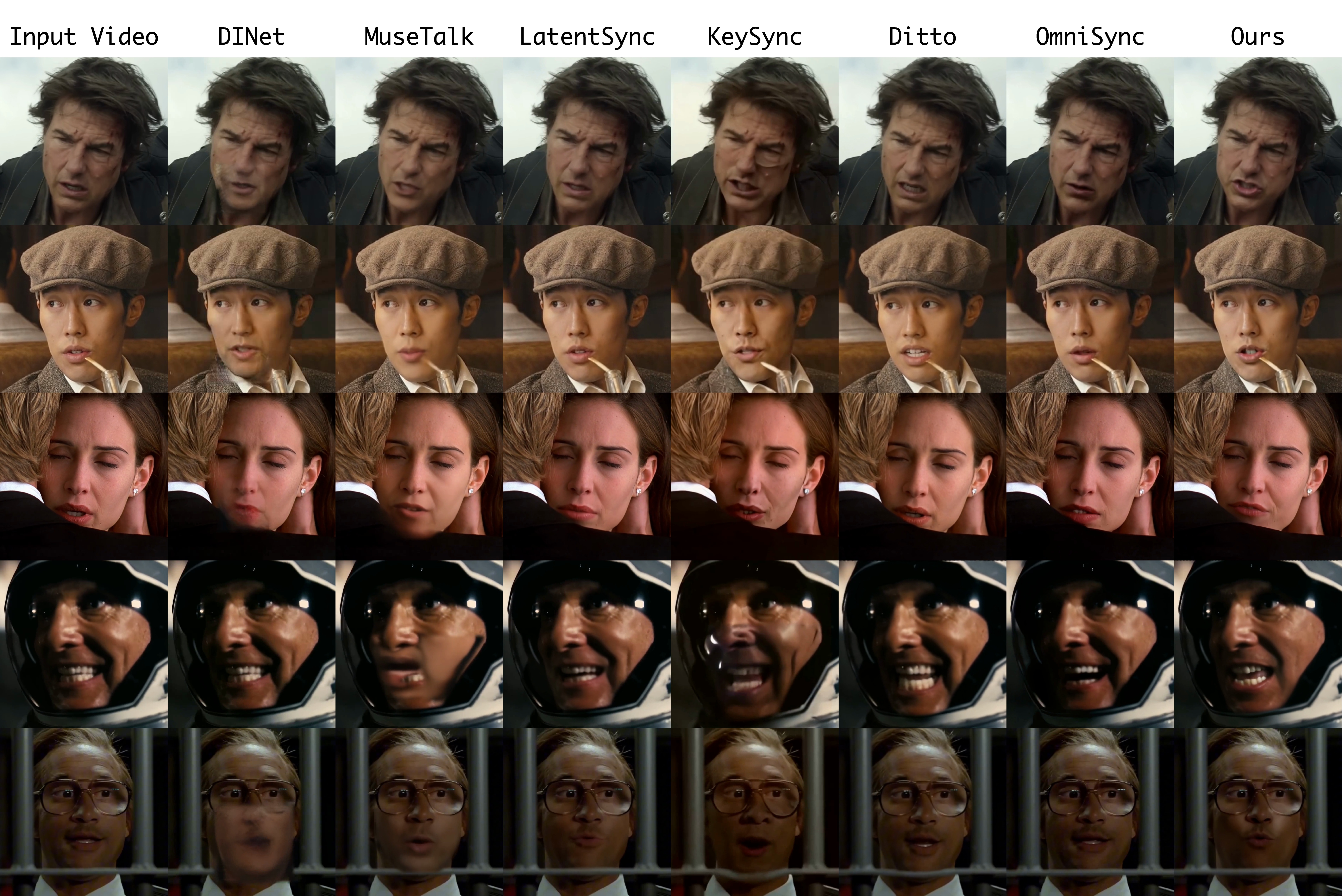}
\caption{Qualitative comparison with existing methods. Refer to Appendix for details. We provide supplementary videos to capture temporal attributes (synchronization, naturalness, stability) missing in static images.}
\label{fig:qualitative_comparison}
\vskip -0.1in
\end{figure}

The first term of \cref{eq:loss_ra}, $\left\| G \left( f_{V_{ab}} \right) - G \left( f_{\hat{V}_{ab}} \right) \right\|_1$, primarily enforces similarity in non-mouth regions between the ground-truth video and the generation conditioned on the correct audio. This encourages the model to faithfully reproduce scene-level factors including occlusions and lighting conditions. The second term, $\left\| G \left( f_{V_{ab}} \right) - G \left( f_{\hat{V}_{cd}} \right) \right\|_1$, captures deviations in non-mouth regions as well as mouth-shape discrepancies caused by mismatched audio, thereby penalizing spurious content leakage from irrelevant acoustic signals. The third term, $\left\| G \left( f_{\hat{V}_{ab}} \right) - G \left( f_{\hat{V}_{cd}} \right) \right\|_1$, isolates the lip-motion difference driven solely by the audio condition. By jointly minimizing the first two terms and preserving the magnitude of the third, our loss effectively decouples lip-sync accuracy from background realism. This ensures that the model generates distinct and accurate mouth movements for different audio inputs while simultaneously maintaining high-fidelity textures in non-mouth regions, including complex occlusions and illumination variations. We further demonstrate this effectiveness through feature-based attention heatmaps, illustrating how our objective enhances structural consistency without compromising audio-driven articulation. 
\section{Experiment}
\subsection{Experimental Settings}
\label{subsec:4_1}
\textbf{Datasets.} We train ComplexSync on a large-scale mixture of datasets, including SpeakerVid-5M~\cite{zhang2025speakervid}, TalkVid~\cite{chen2025talkvid}, and an in-house talking-head video dataset. To ensure high data quality, we apply automated filtering tools such as Koala-36M~\cite{wang2025koala} to remove videos with low brightness or poor aesthetic quality. This standardized curation pipeline yields over 1.2 million high-quality training samples, amounting to more than 2,400 hours of speech-aligned audiovisual content. For evaluation, we use both the HDTF~\cite{zhang2021flow} benchmark and our newly introduced ComplexSync-Val dataset to assess animation fidelity and robustness. Recognizing that existing benchmarks are limited in scope and predominantly feature constrained talking-head scenarios with frontal views and static illumination, we introduce ComplexSync-Val to better reflect real-world challenges. It contains over 200 videos, comprising a balanced mix of real-world recordings and AI-generated content. It features a wide array of demanding conditions, including physical occlusions, dynamic backgrounds, and intricate lighting variations, and is specifically designed to assess the robustness of generative models in unconstrained, open-domain scenes.

\begin{table}[!b]
    \vskip -0.1in
  \centering
  \belowrulesep=0pt
  \aboverulesep=0pt
  {\vskip -1.5mm}
  \renewcommand{\arraystretch}{1.2}
  \scalebox{0.8}{
  \begin{tabular}{l|cccc}
  \toprule
  Methods & Lip-Sync$\uparrow$ & Video Definition$\uparrow$ & Naturalness$\uparrow$ & VA$\uparrow$ \\
  \hline
  DINet & 1.750 & 1.679 & 1.696 & 1.732 \\
  Musetalk & 2.089 & 1.982 & 2.571 & 2.339 \\
  LatentSync & 2.446 & 2.786 & 2.875 & 2.857 \\
  KeySync & 3.393 & 3.196 & 2.696 & 2.857 \\
  Ditto & 3.214 & 3.429 & \underline{3.375} & \underline{3.429} \\
  OmniSync & \underline{3.554} & \underline{3.768} & 3.143 & 3.232 \\
  \hline
  Ours & \textbf{4.239} & \textbf{4.128} & \textbf{4.312} & \textbf{4.129} \\
  \bottomrule
  \end{tabular}
  }
  \vskip -0.05in
  \caption{User Study results. The best results are in \textbf{bold}, and the second-best are in \underline{underlined}.}
  \label{tab:user_study}
  {\vskip -1mm}
\end{table}

\textbf{Implementation Details.} We train our model at $512 \times 512$ resolution using the AdamW optimizer with a fixed learning rate of $1 \times 10^{-5}$. Training proceeds in three stages: in the first stage (\cref{subsec:3_1}), we train for 200K steps on 8 NVIDIA A100 GPUs with a per-GPU batch size of 2; in the second stage (\cref{subsec:3_2}), we train for 50K steps on 4 A100 GPUs with a per-GPU batch size of 4; and in the final stage (\cref{subsec:3_3}), we train for 10K steps using the same setup as the second stage. 

\textbf{Evaluation Metrics.} We evaluate our method using a suite of established metrics: Fréchet Inception Distance (FID) for per-frame visual fidelity, Fréchet Video Distance (FVD) for temporal coherence, CSIM (Cosine Similarity of Identity) for identity preservation, and Sync-C/Sync-D for audio-visual synchronization. To specifically address the challenge of spatial information leakage, we incorporate the LipLeak~\cite{bigata2025keysync} metric introduced by KeySync, which serves as a specialized diagnostic tool to measure a model's susceptibility to identity-induced lip movement leakage. Notably, as ComplexSync-Val includes AI-generated content and cinematic clips lacking ground-truth audio-lip alignment, FID and FVD are excluded from this benchmark. To further evaluate robustness in unconstrained scenarios, we introduce two specialized metrics. First, VFM Feature Distance (VFM-FD) measures fidelity to complex scene factors by computing the $\ell_1$ distance between spatiotemporal features extracted via the frozen VFM described in \cref{subsec:3_3}, analogous to the first term in \cref{eq:loss_ra}. Second, Occlusion Recovery Capability (Occ RC) evaluates occlusion handling by calculating the Intersection-over-Union (IoU) between ground-truth occlusion masks and those extracted from synthesized frames, both obtained via SAM3~\cite{carion2025sam}.

\begin{table}[t!]
    \vskip -0.15in
    \centering
    \belowrulesep=0pt
    \aboverulesep=0pt
    {\vskip 0.8mm}
    \renewcommand{\arraystretch}{1.3}
    \scalebox{0.55}{
    \begin{tabular}{c|c|cccccc}
    \toprule
    Datasets & Methods & FID$\downarrow$ & FVD$\downarrow$ & Sync-C$\uparrow$ & Sync-D$\downarrow$ & LipLeak$\downarrow$ & Occ-RC$\uparrow$ \\
    \hline
    \multirow{7}{*}{HDTF} & w/o Dual-Stream & 7.92 & 40.73 & 4.95 & 9.35 & 0.0246 & - \\
    ~ & \makecell[c]{w/o weight fusion\\(Sync Stream)} & 10.05 & 59.94 & 8.18 & 6.71 & 0.0110 & - \\
    ~ & \makecell[c]{w/o weight fusion\\(Fidelity Stream)} & 8.04 & 119.48 & 1.18 & 13.83 & 0.3116 & - \\
    ~ & w/ Dual-Stream & 7.89 & 36.73 & 6.95 & 7.57 & 0.0153 & - \\
    \cline{2-8}
    ~ & w/o EMA anchor & 8.69 & 53.38 & 6.08 & 8.52 & 0.0698 & - \\
    ~ & DMD2 distillation & 8.62 & 39.45 & 5.42 & 9.03 & 0.1144 & - \\
    ~ & w/ EMA anchor & 8.11 & 38.56 & 7.85 & 6.81 & 0.0103 & - \\
    \hline
    \multirow{3}{*}{\makecell[c]{ComplexSync\\Val}} & w/o RA loss & 10.43 & 66.60 & 9.07 & 6.24 & - & 82.70 \\
    ~ & w/o Gram in RA & 5.48 & 35.92 & 1.69 & 11.94 & - & 98.90 \\
    ~ & w/ Gram in RA & 8.49 & 41.88 & 8.87 & 6.28 & - & 95.20 \\
    \bottomrule
    \end{tabular}
    }
    \vskip -0.1in
    \caption{Quantitative ablation study. Rows 1-4 evaluate dual-stream strategy (w/o Stages 2-3). Rows 5-7, starting from dual-stream, compare distillation schemes to verify our EMA anchor. Final three rows validate RA loss and Gram operation.}
    \label{tab:ab_quantitative_comparison}
\end{table}
\begin{figure}[t]
\centering
\includegraphics[width=0.48\textwidth]{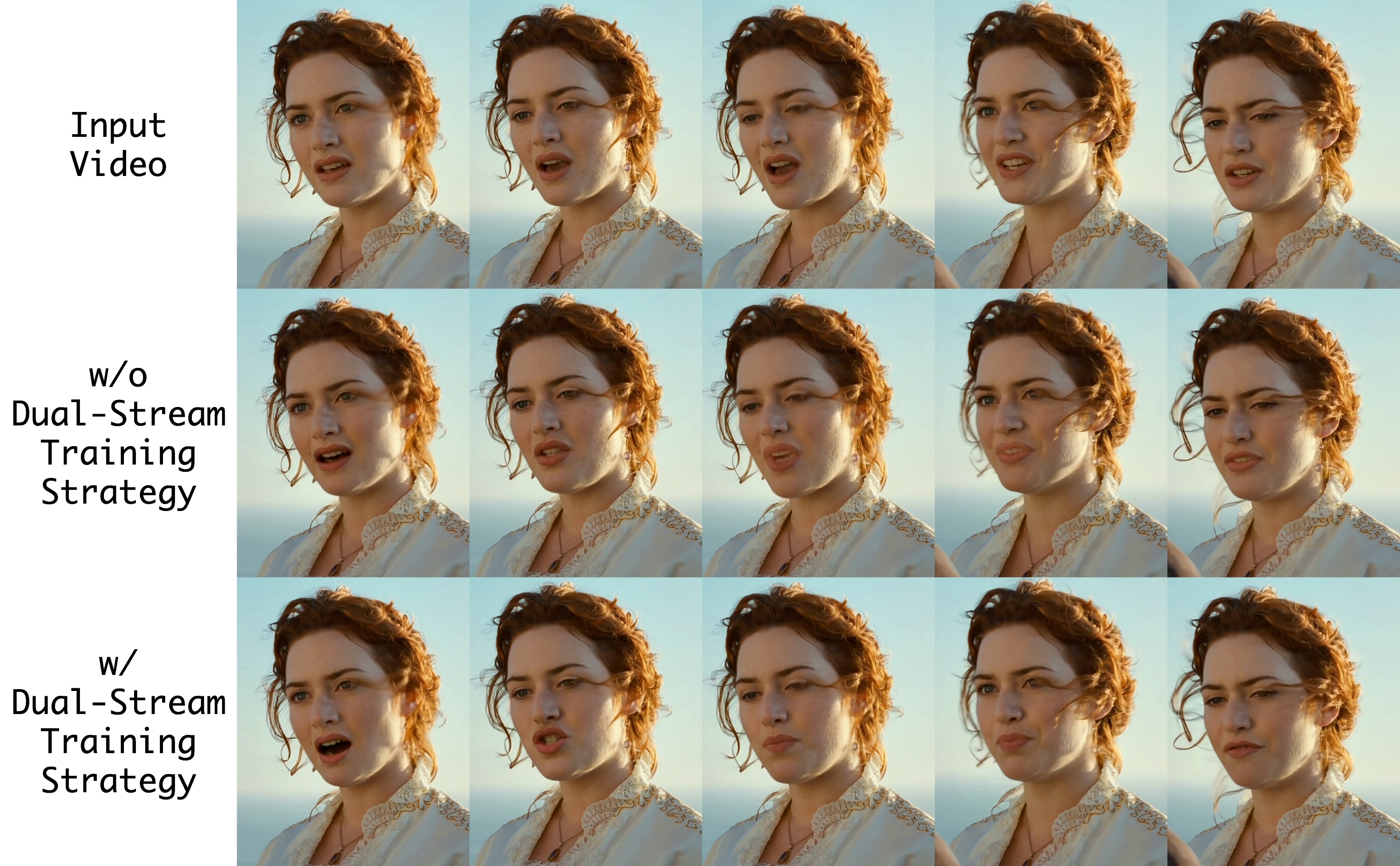}
\caption{Ablation Study for the Dual-stream Joint Training Strategy.}
\label{fig:ab_1}
\vskip -0.2in
\end{figure}

\subsection{Results and Analysis}
\label{subsec:4_2}
To evaluate ComplexSync, we perform extensive comparisons with SOTA baselines such as DINet~\cite{zhang2023dinet}, MuseTalk~\cite{zhang2024musetalk}, Ditto~\cite{li2025ditto}, LatentSync~\cite{li2024latentsync}, KeySync~\cite{bigata2025keysync}, and OmniSync~\cite{pengomnisync}. For OmniSync, since the official implementation remains unavailable, we utilized its commercial interface (Kling) for generating inference outputs for our qualitative and quantitative analysis.

\textbf{Quantitative Results.} As shown in \cref{tab:quantitative_comparison}, ComplexSync achieves the best results across all evaluation metrics on both test datasets. Specifically, it significantly outperforms SOTA methods in visual quality (FID, FVD), identity preservation (CSIM), and lip-sync accuracy (Sync-C, Sync-D). Notably, our method demonstrates remarkable robustness in complex scenarios, evidenced by substantial improvements in LipLeak (0.0213 compared to the second-best 0.0596) and occlusion recovery (Occ-RC). Furthermore, ComplexSync is highly efficient, achieving the fastest inference speed at 72.73 FPS, establishing a new optimal trade-off between generation quality and real-time performance. 

\textbf{Qualitative Results.} We present qualitative comparisons between ComplexSync and existing methods in \cref{fig:qualitative_comparison}. Our approach generates more natural facial expressions and achieves superior lip synchronization. In contrast, MuseTalk, DINet, KeySync, and Ditto exhibit significant visual artifacts and struggle with occlusion recovery. Furthermore, LatentSync and OmniSync often fail to achieve accurate lip shapes, which is likely attributable to inherent information leakage and leads to suboptimal synchronization. Conversely, ComplexSync generates vivid, expressive lip motions and demonstrates exceptional robustness to heavy occlusions, consistently outperforming existing solutions in visual fidelity. For additional comparative details, please refer to the Appendix. Given that static images cannot fully capture critical temporal attributes such as synchronization, naturalness, and stability, we provide comprehensive video comparisons in the supplementary materials.

\textbf{User Study.} To further validate the effectiveness of ComplexSync, we conducted a comprehensive user study involving 50 participants. Users were tasked with evaluating 29 sets of videos, covering both standard scenarios and challenging conditions, using a 5-point Mean Opinion Score (MOS) scale. The evaluation spanned four critical dimensions: Lip Synchronization, Video Definition, Naturalness, and Visual Appeal. As summarized in~\cref{tab:user_study}, ComplexSync consistently outperforms all competing methods across all metrics. This holistic evaluation underscores the superiority of our approach in generating realistic and expressive talking-head animations while maintaining robust identity consistency and high visual fidelity.

\begin{figure}[!t]
\vskip -0.2in
\centering
\includegraphics[width=0.48\textwidth]{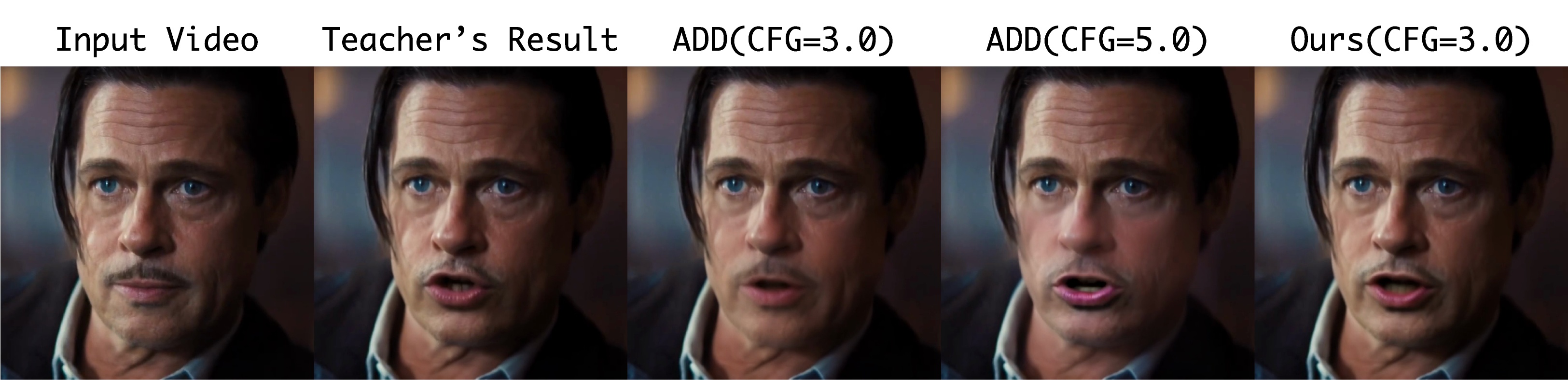}
\vskip -0.1in
\caption{Ablation Study for the Proposed Distillation Strategy.}
\label{fig:ab_2}
\vskip -0.2in
\end{figure}

\subsection{Ablation Study}
\label{subsec:4_3}
\begin{figure*}[!t]
    \vskip -0.1in
\centering
\includegraphics[width=0.9\textwidth]{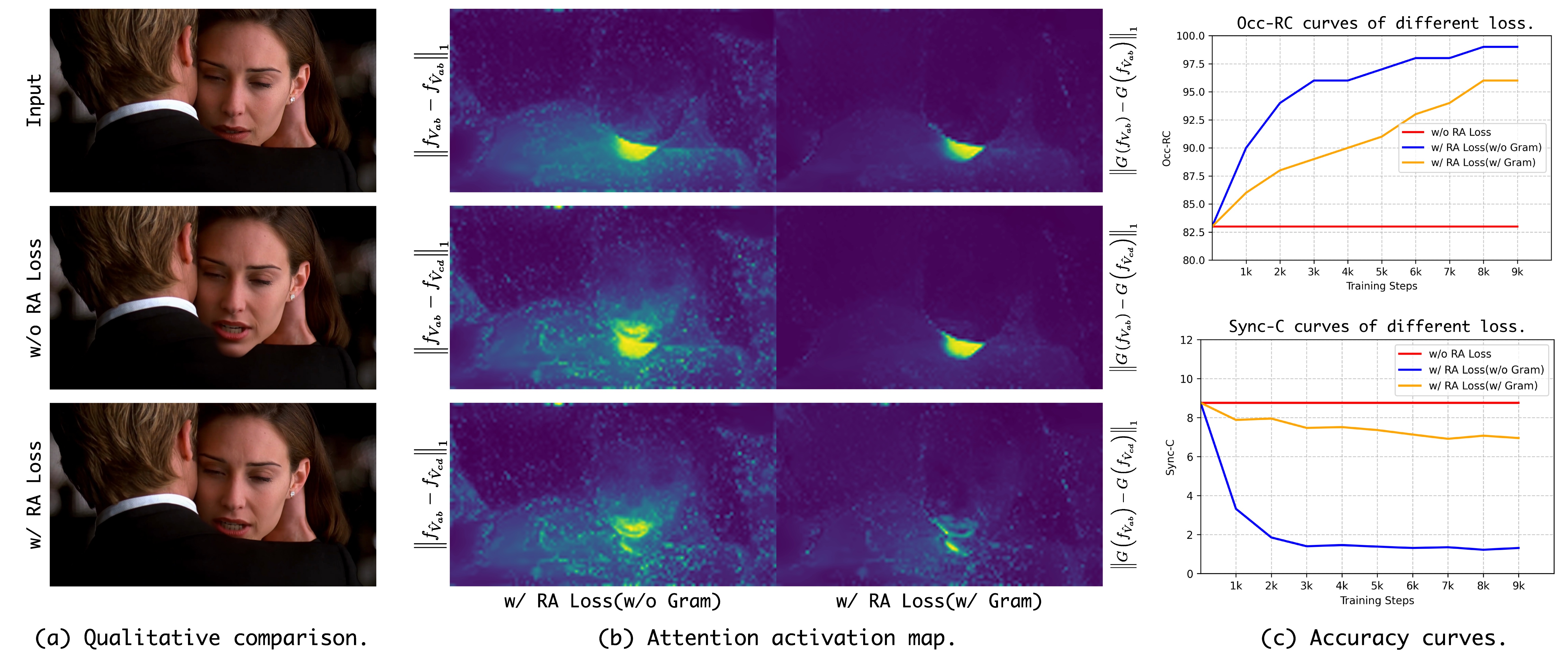}
\vskip -0.1in
\caption{Ablation Study for the Relational Alignment Loss.}
\label{fig:ab_3}
\vskip -0.1in
\end{figure*}

We conduct a comprehensive ablation study on the three core modules of ComplexSync to evaluate their individual contributions. Quantitative ablation results are presented in \cref{tab:ab_quantitative_comparison}, while qualitative ablation results are shown in \cref{fig:ab_1,fig:ab_2,fig:ab_3}.

\textbf{Ablation Study for the Dual-stream Joint Training Strategy.} As shown in rows 1--4 of \cref{tab:ab_quantitative_comparison}, we train several model variants: a single-stream model following the training pipeline of LatentSync and KeySync, a standalone Sync Stream model, a standalone Fidelity Stream model, and our full dual-stream model. Quantitative results demonstrate that dual-stream training effectively balances lip-sync accuracy and visual fidelity, whereas single-stream training suffers from inherent bias (e.g., Fidelity Stream achieves a favorable FID score but yields poor Sync-C performance). As illustrated in \cref{fig:ab_1}, the absence of our dual-stream strategy leads to significant feature leakage, frequently resulting in inaccurate lip synchronization. Furthermore, the generated teeth are noticeably biased by the reference frames, even exhibiting direct replication of the reference tooth textures (e.g., the 3rd and 4th columns of the 2nd row). In contrast, our proposed method effectively resolves these issues by maintaining an optimal balance between audio and visual cues.

\textbf{Ablation Study for the Proposed Distillation Strategy.} Rows 5--7 of \cref{tab:ab_quantitative_comparison} compare different distillation methods, all initialized from the same pretrained model. Quantitative results demonstrate that our distillation approach maintains model performance while significantly reducing inference latency. \cref{fig:ab_2} presents a qualitative comparison between our proposed distillation strategy and the ADD method. As illustrated, distillation via ADD tends to yield suboptimal results across varying CFG scales. Specifically, at a low CFG scale (CFG=3.0), ADD suffers from feature leakage, resulting in failed lip editing and blurred textures; conversely, at a higher CFG scale (CFG=5.0), it encounters identity shift issues. In contrast, our proposed strategy enables high-fidelity generation in a single denoising step while strictly preserving identity consistency and lip-syncing accuracy.

\textbf{Ablation Study for the Relational Alignment Loss.} The last three rows of \cref{tab:ab_quantitative_comparison} validate the RA loss and Gram matrix operation. Quantitative results indicate that removing the Gram matrix yields a higher Occ-RC score but degrades Sync-C due to information leakage (i.e., direct replication of reference frames). (Note: Occ-RC is omitted for HDTF due to absence of occlusions; LipLeak is excluded for ComplexSync-Val due to missing ground truth.) \cref{fig:ab_3} further evaluates individual contributions of the RA loss and Gram matrix. As shown in \cref{fig:ab_3}(a), the model without RA loss fails to reconstruct occluded regions, producing pronounced facial artifacts when objects obscure the face. Moreover, \cref{fig:ab_3}(b) demonstrates that omitting the Gram matrix causes unintended attention biases in non-facial regions, undermining training stability and preventing the loss from functioning as intended. Similarly, \cref{fig:ab_3}(c) highlights that absence of either component precludes achieving optimal balance between lip-syncing accuracy and occlusion recovery. In contrast, our complete method exhibits superior robustness in complex scenarios while maintaining precise equilibrium between driving performance and facial reconstruction.
\section{Conclusion}
We present ComplexSync, a unified framework achieving high-fidelity, real-time lip synchronization in complex scenarios. To mitigate identity leakage from reference frames while preserving natural lip dynamics, we introduce a dual-stream joint training strategy. To overcome the prohibitive latency of iterative diffusion sampling, we develop a novel distillation-based acceleration scheme enabling high-quality single-step denoising at over 70 FPS. We further propose a relational alignment loss that leverages structural priors from VFMs to improve robustness against challenging factors such as occlusions and dynamic lighting. To bridge the evaluation gap, we contribute ComplexSync-Val, the first specialized benchmark for complex lip synchronization, which comprises over 200 diverse video sequences and specialized metrics. Experiments validate that ComplexSync achieves state-of-the-art performance across both standard and complex scenarios, establishing a new frontier for efficient and robust lip sync.
{
    \small
    \bibliographystyle{ieeenat_fullname}
    \bibliography{main}
}

\end{document}